\documentclass[10pt,twocolumn,letterpaper]{article}
\usepackage{authblk}
\usepackage[symbol]{footmisc}
\usepackage[misc]{ifsym}
\usepackage{graphicx}

\makeatletter
\renewcommand\AB@affilsepx{\quad}
\makeatother
\usepackage[pagenumbers]{cvpr} % To force page numbers, e.g. for an arXiv version

\definecolor{cvprblue}{rgb}{0.21,0.49,0.74}
\usepackage[pagebackref,breaklinks,colorlinks,allcolors=cvprblue]{hyperref}

\def\paperID{CVEU-18} % *** Enter the Paper ID here
\def\confName{CVPR}
\def\confYear{2025}

\title{Generative AI for Film Creation: A Survey of Recent Advances}

\author{
   Ruihan Zhang\textsuperscript{1,*,$\dagger$\Letter},
   Borou Yu\textsuperscript{2,*,$\dagger$\Letter},
   Jiajian Min\textsuperscript{3,*},\\
   \vspace{-11pt}
   Yetong Xin\textsuperscript{4},
   Zheng Wei\textsuperscript{17},
   Juncheng Nemo Shi\textsuperscript{5},
   Mingzhen Huang\textsuperscript{6},
   Xianghao Kong\textsuperscript{17},
   Nix Liu Xin\textsuperscript{7},
   Shanshan Jiang\textsuperscript{8},
   Praagya Bahuguna\textsuperscript{9},
   Mark Chan\textsuperscript{9},
   Khushi Hora\textsuperscript{9,16},
   Lijian Yang\textsuperscript{10},
   Yongqi Liang\textsuperscript{10},
   Runhe Bian\textsuperscript{4},
   Yunlei Liu\textsuperscript{11},
   Isabela Campillo Valencia\textsuperscript{12},
   Patricia Morales Tredinick\textsuperscript{13},
   Ilia Kozlov\textsuperscript{14},
   Sijia Jiang\textsuperscript{4},
   Peiwen Huang\textsuperscript{4},
   Na Chen\textsuperscript{15},
   Xuanxuan Liu\textsuperscript{4},\\
   Anyi Rao\textsuperscript{17,$\dagger$\Letter}
\\
\vspace{15pt}
\textsuperscript{1}{Google}
\textsuperscript{2}{University of California, Santa Barbara, Media Arts and Technology}
\textsuperscript{3}{MYStudio}\\
\textsuperscript{4}{Harvard University}
\textsuperscript{5}{Reality Hack}
\textsuperscript{6}{SUNY Buffalo}
\textsuperscript{7}{Onceness}
\textsuperscript{8}{University of Southampton}\\
\textsuperscript{9}{New York University}
\textsuperscript{10}{Communication University of China}
\textsuperscript{11}{University of Southern California}
\textsuperscript{12}{Dodge College of Film and Media Arts}
\textsuperscript{13}{Pratt Institute}
\textsuperscript{14}{Rubyspot}
\textsuperscript{15}{MIT}
\textsuperscript{16}{Netflix}\\
\textsuperscript{17}{Hong Kong University of Science and Technology}
 }

\begin{document}
\maketitle
% !TEX root = ../main.tex
\begin{abstract}
Generative AI (GenAI) is transforming filmmaking, equipping artists with tools like text-to-image and image-to-video diffusion, neural radiance fields, avatar generation, and 3D synthesis. This paper examines the adoption of these technologies in filmmaking, analyzing workflows from recent AI-driven films to understand how GenAI contributes to character creation, aesthetic styling, and narration. We explore key strategies for maintaining character consistency, achieving stylistic coherence, and ensuring motion continuity. Additionally, we highlight emerging trends such as the growing use of 3D generation and the integration of real footage with AI-generated elements.

Beyond technical advancements, we examine how GenAI is enabling new artistic expressions, from generating hard-to-shoot footage to dreamlike diffusion-based morphing effects, abstract visuals, and unworldly objects. We also gather artists' feedback on challenges and desired improvements, including consistency, controllability, fine-grained editing, and motion refinement. Our study provides insights into the evolving intersection of AI and filmmaking, offering a roadmap for researchers and artists navigating this rapidly expanding field.

\end{abstract}
% \footnotetext[]{\dag These authors contributed equally to this work.} 
% \footnotetext[]{\textsuperscript{*}Corresponding author}   
\footnotetext[1]{These authors contributed equally to this work.}
\footnotetext[2]{Corresponding author: anyirao@ust.hk, ruihanz@google.com, anna.yu@aya.yale.edu}
% !TEX root = ../main.tex
\section{Introduction}
% \label{sec:intro} @ruihan

% To insert a figure: \input{figs/template}
% Or table: \input{tables/template}

In recent years, generative AI (GenAI) has made significant advances in video generation with diffusion models~\cite{liu2024sora,yariv2024diverse,yang2024cogvideox,ren2024customize, guo2023animatediff}, 3D asset creation with Gaussian Splatting and NeRF-based models~\cite{kerbl3Dgaussians,mildenhall2021nerf,yao2024neural,li2024advances,bahmani2024tc4d,petrov2024gem3d, mu2024gsdiffusion}, and avatar synthesis~\cite{wang2023rodin, luo2021normalized}. AI-driven content creation is becoming increasingly powerful, enabling AI filmmaking. Over the past few years, we have witnessed a growing number of AI-generated films.

However, the artistic and academic communities remain largely disconnected. Artists often lack insight into where stochasticity originates, e.g. why maintaining character consistency is so difficult, why descriptions of multiple characters in the same frame can lead to confusion. Conversely, researchers have little knowledge of effective artistic workflows and creative needs—do artists truly require one-minute-long generated clips? To what extent do they need controllability over character and camera movement?
%real-world
%, and why generating drone views is more challenging than close-up shots
%Which features should be prioritized for development? What aesthetic standards do they seek? 

In this paper, we analyze user survey data from the MIT AI Film Hack, a filmmaking hackathon that has collected hundreds of AI films over three years (2023\cite{hack_2023}, 2024\cite{hack_2024}, 2025\cite{hack_2025}). We examine the adoption rates of various GenAI tools based on the submission data. We also conduct a quantitative analysis of artists' concerns and expectations, and present case studies showcasing how artists efficiently utilize these tools in their creative workflows. We aim to provide a comprehensive overview of the AI filmmaking landscape, offering insights into current trends, best practices, and key challenges in AI filmmaking.

\section{Related Work}
\label{sec:related}

\subsection{Understanding Film Production}
Traditionally, film production consists of three phases: pre-production (scriptwriting, storyboarding, character design), production (direction, cinematography \cite{xu2025transforming,wei2023feeling,xu2023cinematography}, and other departments working in tandem \cite{wei2024hearing,wei2024multi}), and post-production (editing, special effects, sound design, and mixing) \cite{honthaner2013complete,wei2025illuminating}. With advances in multimodal AI, researchers have developed text-to-video models that generate coherent clips or full AI-driven shorts from minimal input, synthesizing both appearance and motion \cite{liu2024sora,yariv2024diverse,yang2024cogvideox,ren2024customize}. In recent multimodal generative models research, certain pipelines demonstrate the ability to maintain internal continuity, style coherence, and believable character interplay across challenging scene transitions \cite{zhou2025storydiffusion,zhu2024zero,zhang2023adding,jiang2024autonomous}. Beyond 2D frames, AI extends into 3D and 4D asset generation, where virtual actors integrate high-resohere models such as Neural Radiance Fields (NeRF), 3D Gaussian Splatting and dynamic 3D representations enable realistic scene synthesis and temporal consistency \cite{kerbl3Dgaussians,mildenhall2021nerf,yao2024neural,li2024advances,bahmani2024tc4d,petrov2024gem3d}. These approaches allow filmmakers to generate and animate objects and environments over time, ensuring spatial and stylistic fidelity \cite{wu2025automatic}. Another critical advancement is AI-driven avatar creation and human motion synthesis, where digital characters are synthesized using neural geometry and motion priors, producing full-body avatars with expressive facial features and physical plausible movement \cite{tevet2023human,li2024favor,qiu2024moviecharacter,zhu2024trihuman}. These virtual actors integrate high-resolution textures and skeleton rigs to deliver nuanced performances, reducing the need for large-scale casting or motion capture \cite{liu2025make,liao2024tada}. AI-driven avatars also offer style adaptability, allowing seamless transitions between photorealistic and artistic aesthetics without asset reconstruction \cite{zhou2024ultravatar}. By automating complex tasks across pre-production, production, and post-production, AI is reshaping the filmmaking pipeline, reducing costs, streamlining workflows, and enhancing creative control, ultimately enabling richer storytelling with fewer logistical constraints.

%Ref: Survey
%Video Gen, 3D, Avatar

\subsection{AI Film Workflow}
The earliest AI–movie integrations primarily targeted perception tasks—scene analysis, object detection, or camera calibration—to partially automate editing and tagging \cite{rao2020local,xia2020online,rao2020unified,rao2022shoot360,rao2022coarse}. Today, AI spans all production phases to elevate both creativity and efficiency \cite{rao2024scriptviz,jiang2024cinematic,guo2024sparsectrl,zhang2023adding,ma2023automated,rao2023dynamic}. In pre-production, generative language models help refine scripts and produce initial concept art \cite{menon2025revolutionizing,mirowski2023co,muller2022genaichi,kim2024unlocking,mahon2024screenwriter}. Meanwhile, diffusion-based storyboard generation can block out potential shots—complete with lighting or basic character poses—by interpreting textual scene descriptions. During production, real-time vision algorithms perform automated camera positioning, track actors’ locations, or match composite elements (like CGI props) to real set positions \cite{he2023virtual,nageli2017real,stoll2023automatic,wu2025automatic}. Techniques can also incorporate generative volumetric backdrops, instantly turning minimal green-screen footage into richly detailed sets \cite{mildenhall2021nerf,martin2021nerf,li2022neural}. Finally, in the post-production phase, automated shot segmentation, character detection, and special-effects composition substantially reduce the labor intensity of traditional editing workflows \cite{xia2020online,zhu2023autoshot,stoll2023automatic,rao2022coarse}. Meanwhile, large-scale datasets like MovieNet~\cite{huang2020movienet} provide multi-modal annotations that establish standardized benchmarks for shot composition, narrative structure, and affective analysis \cite{song2024moviellm}. To enhance emotional impact, researchers focus on the three main media components—visuals \cite{zheng2024open,xing2024make,zhang2023adding}, sound \cite{tian2024vidmuse}, and editing \cite{argaw2022anatomy,zhu2023moviefactory}—and apply AI to music generation \cite{civit2022systematic,kang2024video2music,xue2025audio} and automated editing \cite{wang2023reprompt}. Moreover, classical narrative frameworks such as the Hero’s Journey \cite{wei2024telling} and the Freytag’s Pyramid \cite{yang2021design} are now informed by deep learning models that offer automated analysis and generative support for script pacing and character development \cite{mirowski2023co}. Overall, as generative models and multi-modal learning continue to advance, AI’s role in filmmaking—spanning creation, stylistic coherence, and audience engagement—will likely deepen, creating new opportunities and challenges for the industry.

%\subsection{AI film workflow}

%(Jiajian reference)
%pre(script stroyboard), 
%production, 
%post production (vfx) 

% !TEX root = ../main.tex
\section{Survey}
\label{sec:survey}

% https://docs.google.com/spreadsheets/d/1D9zcJEPO4-fzskkRT3luhfuBz5lxivPcEW6b2I6v2Ns/edit?gid=909563654#gid=909563654

% https://docs.google.com/spreadsheets/d/1D9zcJEPO4-fzskkRT3luhfuBz5lxivPcEW6b2I6v2Ns/edit?gid=909563654#gid=909563654

% With the rapid advancement of GenAI, 
Many computer scientists and filmmakers are eager to understand how these tools are being integrated into film production. In this section, we examine the adoption rates of various GenAI technologies and how artists rate the impact of different factors on film quality. We also surveyed artists on their opinions of current GenAI tools and their expectations for future advancements in film production.

\begin{table*}[h]
\centering
\begin{tabular}{lccc}
\hline
 \textbf{Category} & \textbf{2023 (n=8)} & \textbf{2024 (n=67)} & \textbf{2025 (n=118)}  \\
\hline
LLM-assisted scriptwriting & 37.5\% & - & 54.2\%  \\ 
AI-generated video & 87.5\% & 95.5\% & 100.0\% \\
AI-generated 3D assets & 0.0\% & 20.9\% & 23.7\% \\
AI-generated voiceovers & 0.0\% & 59.7\% & 53.4\% \\
AI-generated music/sound effects & 12.5\% & 50.7\% & 54.2\% \\
Blending real and AI-generated footage & 12.5\% & 1.5\% & 17.8\% \\
\hline
\end{tabular}
\caption{Adoption rate of GenAI tools in the MIT AI Film Hack from 2023 to 2025. Percentages indicate the proportion of films utilizing each tool in a given year. A '-' signifies that the item was not surveyed in that year. Note that not all films include voiceovers, meaning the actual adoption rate of AI voiceover tools among films with voiceover is higher. }
\label{tab:genai_adoption}
\end{table*}

\subsection{GenAI Tools Adoption Rate}

%scriptwriting, storyboarding, shooting, and post-production, including editing, music, and voiceover
Filmmaking traditionally involves various stages like scriptwriting, video generation, 3D content creation, music composition, and voiceover. We aim to explore how AI can contribute to different stages of this pipeline. We surveyed the use of various GenAI tools in the MIT AI Film Hack~\cite{hack_2023,hack_2024,hack_2025}, an event that challenges participants to create short films using AI. Running annually in 2023~\cite{hack_2023}, 2024~\cite{hack_2024}, and 2025~\cite{hack_2025}, this film hack provides valuable insights into the utilization of AI in various stages (Table \ref{tab:genai_adoption}).

We observed that nearly all participants incorporated image or video generation in the production (Table \ref{tab:genai_adoption}). Despite the increasing realism of AI-generated visuals \cite{openai2024}, most films retained a cartoonish style—likely because inconsistencies are less noticeable than realistic style videos \cite{hack_2025}.

% Seeing the potential of 3D generation in enhancing temporal and spatial consistency, MIT AI Film Hack organizers introduced a dedicated 3D generation track in 2024~\cite{hack_2024} to encourage broader adoption of AI-driven 3D content. 
Recognizing 3D generation’s potential to bolster both temporal and spatial consistency, MIT AI Film Hack organizers introduced a dedicated 3D generation track in 2024 \cite{hack_2024}. This led to a notable increase in 3D tool usage from 0\% in 2023 to 23.7\% in 2025 (Table \ref{tab:genai_adoption}). Yet the adoption rate remains lower than video gen tools (Table \ref{tab:genai_adoption}), suggesting that AI-based 3D generation still faces challenges in meeting filmmakers’ expectations.

By 2024, more than half of the films incorporated AI voiceovers (Table \ref{tab:genai_adoption}). Notably, non-native English speakers particularly embraced AI voiceover tools, using them to generate seamless, natural-sounding English narration for a global audience\cite{tale_of_lipu_village,earth_home_turkey_farm,what_do_you_call_home}, demonstrating how AI enables borderless artistic expression and accessibility.

AI-generated music and sound effects saw increasing adoption, rising from 12.5\% in 2023 to over 50\% in 2024 and 2025 (Table \ref{tab:genai_adoption}). Interestingly, the winning music piece in the 2025 competition was created by a human composer \cite{qatsi}, despite AI's significant presence. Because the judges were blind to AI usage and evaluated solely on the music’s emotional effectiveness, these findings suggest that human-created compositions still hold an advantage in capturing nuanced emotion and variation.

\begin{table}[h]
\centering
\begin{tabular}{ll}
\hline
\textbf{Feature} & \begin{tabular}{@{}c@{}}\textbf{Importance} \\ mean (s.e.)\end{tabular} \\
\hline
Selecting appropriate video gen tools & 6.45 (0.073) \\
Crafting detailed prompts & 5.91 (0.117) \\
Providing detailed style descriptions & 5.65 (0.127) \\
Generating multiple iterations & 6.05 (0.116) \\
\hline
\end{tabular}
\caption{Best practices for using video generation tools surveyed from 110 artists. Importance rated on a scale of 0-7.}
\label{tab:video_factors}
\end{table}

% \begin{table*}[h]
% \centering
% \begin{tabular}{lccc}
% \hline
%  \textbf{Category} & \begin{tabular}{@{}c@{}}\textbf{2023} \\ n=8 \end{tabular}  & \begin{tabular}{@{}c@{}}\textbf{2024} \\ n=67 \end{tabular}  & \begin{tabular}{@{}c@{}}\textbf{2025} \\ n=118 \end{tabular}  \\
% \hline
% LLM co-write script & 3/8 & - & 64/118  \\ 
% Video generation  & 7/8 & 64/67 & 118/118 \\
% 3D asset generation & 0/8 & 14/67 & 28/118 \\
% AI Voiceover & 0/8 & 40/67 & 63/118 \\
% AI music/sound effect & 1/8 & 34/67 & 64/118 \\
% Combine real and AI footage & 1/8 & 1/67 & 21/118 \\
% \hline
% \end{tabular}
% \caption{Adoption rate of GenAI tools in MIT AI Film Hack 2025.}
% \label{tab:genai_adoption}
% \end{table*}

\subsection{GenAI Tools and Film Quality}

Through surveys of experienced AI film creators, we gathered insights into user preferences on key aspects of video generation (Table \ref{tab:video_factors}) and 3D generation tools (Table \ref{tab:3d_factors}).

People regarded selecting the right GenAI tools as the most critical factor in achieving high-quality films. The competitive landscape is evident in the diverse range of tools reported by users, including Midjourney~\cite{midjourney}, Kling~\cite{klingai}, OpenArt~\cite{openart}, Runway~\cite{runway}, and Pixverse~\cite{pixverse}. Notably, artists in the 2025 MIT AI Film Hack used an average of three tools per film (Table \ref{tab:number_of_tools}), suggesting that different tools play complementary roles in meeting visual expectations, as no single tool fully replaces the others on the market.

Survey responses emphasized the need for multiple generation iterations to mitigate the impact of stochasticity in individual AI-generated outputs (Tables \ref{tab:video_factors} and \ref{tab:3d_factors}). Additionally, users highlighted the importance of crafting detailed prompts, often utilizing prompt rewriting tools, to achieve visually rich and appealing results in both 2D and 3D generation tasks (Tables \ref{tab:video_factors} and \ref{tab:3d_factors}).

\begin{table}[ht]
\centering
\begin{tabular}{lc}
\hline
\textbf{Feature} & \begin{tabular}{@{}c@{}}\textbf{Number of tools per film} \\ mean (s.e.)\end{tabular} \\
\hline
MIT AI Film Hack 2023 & 2.50 (0.327) \\
MIT AI Film Hack 2024 & 3.46 (0.210) \\
MIT AI Film Hack 2025 & 3.14 (0.136) \\ %3.144067797(0.1357854823)\\
\hline
\end{tabular}
\caption{Number of video gen tools used in one film in the MIT AI Film Hack 2023, 2024 and 2025.}
\label{tab:number_of_tools}
\end{table}

\begin{table}[h]
\centering
\begin{tabular}{ll}
\hline
\textbf{Feature} & \begin{tabular}{@{}c@{}}\textbf{Importance} \\ mean (s.e.)\end{tabular}  \\
\hline
Choosing the right genAI product & 5.97 (0.118)\\
Writing detailed prompts & 5.33 (0.160)  \\
Trying the generation multiple times & 5.34 (0.173)\\
\hline
\end{tabular}
\caption{Best practices for using 3D generation tools (n=65). Importance rated on a scale of 0-7.}
\label{tab:3d_factors}
\end{table}

\subsection{Artists’ Expectations for GenAI Tools}

% We observed significant challenges in maintaining character consistency and generating smooth movements. Additionally, achieving accurate camera angles and framing—especially in wide shots—proved difficult, likely due to biases in training data related to varied viewpoints and focal lengths.

% Analysis of artist surveys (Table \ref{tab:video_expectation}) reveals that consistent character movement is the highest-priority feature for video generation tools, followed by camera control and overall character consistency. This aligns with artists’ expectations for GenAI outputs to exhibit naturalness and coherence comparable to traditional filmmaking, as real-world footage inherently maintains spatial-temporal consistency due to physics.
% While user practices play a crucial role, the fundamental determinant of perceived video quality remains the underlying capabilities of GenAI models. Survey responses from artists (Table \ref{tab:video_expectation}) indicate that consistent character movement is the top priority for video generation tools, followed by camera control and overall character consistency. This finding aligns with artists’ expectations that GenAI outputs should achieve the naturalness and coherence of traditional filmmaking, where real-world footage inherently preserves spatial-temporal consistency due to physics.

While user practices are important, GenAI models’ capabilities ultimately dictate perceived video quality. Survey responses from artists (Table \ref{tab:video_expectation}) rank consistent character movement as the top priority for video generation tools, followed by camera control and overall character consistency. This aligns with artists’ expectations for GenAI outputs to match the naturalness and coherence of real-world filmmaking, where physics inherently enforces spatial-temporal consistency.

% Character consistency has always been a priority for users. In the first iteration of the MIT AI Film Hack in 2023, three out of eight films featured dogs as main characters—a strategic workaround for the limitations of AI-generated human characters~\cite{hack_2023}. Unlike humans, where small facial differences are easily noticeable, dogs tend to appear more similar with simple descriptions, making them a workaround for maintaining character consistency. 

% People express a strong desire for greater control over camera angles. In real-world filmmaking, camera movement follows a complex 3D trajectory, involving not just the degree of motion (e.g., "pan left 0.1–10") but also factors such as the starting position, focal length, and depth of field. Accurately describing these elements can be challenging. Additionally, even with precise descriptions, the model often struggles to generate certain perspectives, such as drone views or long-distance shots  This limitation is likely due to a lack of sufficient training data for these specific viewpoints.
Users seek finer control over camera angles. In real-world filmmaking, camera movement follows a complex 3D trajectory—encompassing factors like starting position, focal length, and depth of field, not just a single motion parameter (e.g., “pan left 0.1–10”). Describing these elements is challenging, and models often struggle with perspectives like drone or long-distance shots due to insufficient training data. 

We also observed a strong desire for controlling character movement (Table \ref{tab:video_expectation}), encompassing both expressive motion within individual clips and consistent movement across camera angle changes as a visual hook. There is also significant interest in generating multiple characters within a single frame (Table \ref{tab:video_expectation}), particularly among users seeking more complex narratives in longer video productions.

% We also observed a strong demand for controlling character movement (Table \ref{tab:video_expectation}). This includes not only controlled movement within individual clips to enhance expressivity, but also ensuring consistent motion across camera angle changes between adjacent clips, serving as a visual hook.

% as movement plays a crucial role in expression. Additionally, maintaining consistent movements is often essential for seamless transitions between adjacent clips, serving as a visual hook.

% Significant interest exists in generating multiple characters within a single frame (Table \ref{tab:video_expectation}), particularly for users seeking more complex narratives in longer video productions.

% we saw som  tools uses motion capture to provide a good refrence like Autodesk, though its requires VR headset for motion capture. Other tools uses skeleton to control body movement but it is inconvenient to manipulate on scren without 3D software.

\begin{table*}[h]
\centering
\resizebox{\textwidth}{!}{
\begin{tabular}{lcc}
\hline
\textbf{Task} &  \textbf{Importance} & \textbf{Current tools performance}\\
\hline
Generate consistent characters according to the reference image/text description & 6.34 (0.103) & 4.45 (0.140) \\
Generate multiple main characters in one frame   & 6.15 (0.100) & 3.91 (0.168) \\
Generate consistent character body movement & 6.62 (0.072) & 4.55 (0.147) \\
Follow the instruction in character body movement & 6.18 (0.103) & 4.06 (0.162) \\
Control camera movement  & 6.35 (0.082) & 4.71 (0.126) \\
Allow local editing         & 6.24 (0.102) & 4.33 (0.155) \\
\hline
\end{tabular}
}
\caption{Artists' expectation for video gen tools for filmmaking (n=100).}
\label{tab:video_expectation}
\end{table*}

% ,  suggesting that 3D GenAI still has significant room for improvement. 
We also surveyed users' ratings of 3D generation tools and found lower satisfaction compared to video generation tools. Many users reported that the generated 3D meshes often lack the desired styles and proper mesh topology (Table \ref{tab:3d_expectation}). Additionally, many 3D generation tools struggle to create fine structures (Table \ref{tab:3d_expectation}), such as hollow designs or intricate details.

\begin{table*}[ht]
\centering
\begin{tabular}{lcc}
\hline
\textbf{Task} &  \textbf{Importance} & \textbf{Current tools performance}\\
\hline
Generate decent meshes & 6.09 (0.127)  & 3.99 (0.165)\\
Generate fine structures & 6.24 (0.115) & 3.84 (0.179)\\
Generate the desired styles  & 6.28 (0.112) & 3.94 (0.166)\\
\hline
\end{tabular}
\caption{Artists' expectation for 3D gen tools for filmmaking (n=65). Artists were asked to rate how well 3D gen tools perform in certain aspects and also how important these features are.}
\label{tab:3d_expectation}
\end{table*}

% \begin{table*}[h]
% \centering
% \begin{tabular}{lll}
% \hline
% \textbf{Task} &  \textbf{Importance} & \textbf{Current tools performance}\\
% \hline
% Generate consistent characters according to the reference image/text description     & 6.336363636(0.1027375684) & 4.454545455(0.1388841289) \\
% Generate multiple main characters in the sample frame   & 6.145454545(0.0999051791) & 3.909090909(0.1681699836) \\
% Generate consistent character body movement     & 6.618181818(0.07188840646) & 4.545454545(0.1470512859) \\
% Follow the instruction in character body movement & 6.181818182(0.103469277) & 4.063636364(0.1624855278) \\
% Control camera movement      & 6.354545455(0.08228022627) & 4.709090909(0.1261117067) \\
% Allow local editing         & 6.236363636(0.1024530428) & 4.327272727(0.1550426189) \\
% \hline
% \end{tabular}
% \caption{Artists expectation for video gen tools for filmmaking}
% \label{tab:video_expectation}
% \end{table*}

% \begin{table*}[h]
% \centering
% \begin{tabular}{lll}
% \hline
% \textbf{Task} &  \textbf{Importance} & \textbf{Current tools performance}\\
% \hline
%  generating decent meshes   & 6.089552239(0.1271248052)  & 3.985074627(0.1654080071)\\
%  generating fine structures   & 6.23880597 (0.1147028336) & 3.835820896(0.1786894029)\\
%  generating the desired style    & 6.28358209(0.1121436416) & 3.940298507(0.1659377053)\\
% \hline
% \end{tabular}
% \caption{Artists expectation for 3D gen tools for filmmaking(n=65)}
% \label{tab:3d_expectation}
% \end{table*}
% !TEX root = ../main.tex
\section{Case Studies}

Generative AI is reshaping filmmaking by enabling novel aesthetic and narrative strategies while demanding new forms of creative control. Through analyzing award-winning projects from the MIT AI Filmmaking Hackathons\cite{hack_2023,hack_2024,hack_2025}, we examine how creators navigate this evolving landscape across domains.

\subsection{Case Study of Visual Storytelling} 

Generative AI enables new forms of visual storytelling by supporting aesthetic consistency, shot composition, and camera movement, though it still relies on human intervention for narrative coherence and stylistic control\cite{zhou2025storydiffusion, guo2023animatediff}.

\subsubsection*{Aesthetic Styling}
% Filmmakers use AI tools in conjunction with traditional techniques to achieve distinctive and coherent visual styles across scenes. Techniques include prompt engineering, image references, LoRA model training\cite{hu2021lora}, and tools like AnimateDiff\cite{guo2023animatediff}. Artists often supplement these techniques with hand-drawings, digital collaging, and post-production to achieve desired effects.

Filmmakers combine AI tools with traditional techniques to achieve distinctive and coherent visual styles across scenes. Techniques include prompt engineering, image referencing, LoRA model training\cite{hu2021lora}, and tools like AnimateDiff\cite{guo2023animatediff}. Artists often enhance these methods with hand-drawings, digital collaging, and post-production to attain desired effects.

\begin{figure}[]
    \centering
     \includegraphics[width=\linewidth]{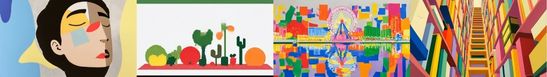}
    \caption{\textit{A Dream About to Awaken}\cite{dream2025} visual style}
    \label{fig:1 A Dream About to Awaken.jpg}
\end{figure}

First, artists can generate coherent styles for AI tools using textual and visual references. For example, by referencing early multiflash photography studies of motion, \textit{O.R.V. 8}\cite{orv8_mit_ai_film_2025} merged historical aesthetics with contemporary palettes. Similarly, \textit{A Dream About to Awaken}\cite{dream2025} leverages prompts derived from AI image interpretation tools applied to hand-drawn storyboards, remixing them with diverse colors and styles to form a unique visual language (Fig \ref{fig:1 A Dream About to Awaken.jpg}).

% Second, hand sketches can serve as a stylistic foundation for genAI tools. \textit{Round Table}\cite{Round_Table} began with hand sketching and digital collaging, using them as guides for AI-generated visuals to ensure consistency. 
Second, hand sketches can serve as a stylistic foundation for generative AI tools, as demonstrated by the film \textit{Round Table} \cite{Round_Table}. There, initial hand sketches and digital collages guided the AI generation process to ensure visual consistency.

% Third, training custom LoRA models\cite{hu2021lora} on curated datasets is another popular approach. In \textit{Overthinking}\cite{Overthinking}, a specialized LoRA model\cite{hu2021lora} trained on 50 mid-century toy images evokes a nostalgic, minimalist style. This short film also incorporates Stable Diffusion's AnimateDiff\cite{guo2023animatediff}, implemented through ComfyUI workflow\cite{comfyui2023} with IPAdapter\cite{liu2023ipadapter}, to maintain stylistic consistency—particularly evident in elements like the chat bubbles.

Third, training custom Low-Rank Adaptation (LoRA) models\cite{hu2021lora} is another popular approach for establishing specific visual styles. For example, the short film \textit{Overthinking}\cite{Overthinking} employed a specialized LoRA model trained on 50 mid-century toy images to evoke a nostalgic, minimalist aesthetic. To maintain this stylistic consistency across animated sequences, particularly evident in elements like chat bubbles, the film also incorporated AnimateDiff\cite{guo2023animatediff}, implemented via a ComfyUI workflow\cite{comfyui2023} with IPAdapter\cite{liu2023ipadapter}.

Finally, traditional post-production techniques remain essential for polishing AI-generated visuals. \textit{Qatsi}\cite{qatsi} integrated AI-generated abstract imagery with film grain and color grading, grounding its ethereal montages in the tactile texture of early cinema. Similarly, the creators of \textit{Round Table}\cite{Round_Table} manually refined and assembled AI-generated assets using standard editing software (Photoshop\cite{adobe_photoshop}, CapCut\cite{capcut}) to impart a handcrafted, tactile feel.

\begin{figure}[]
    \centering
     \includegraphics[width=\linewidth]{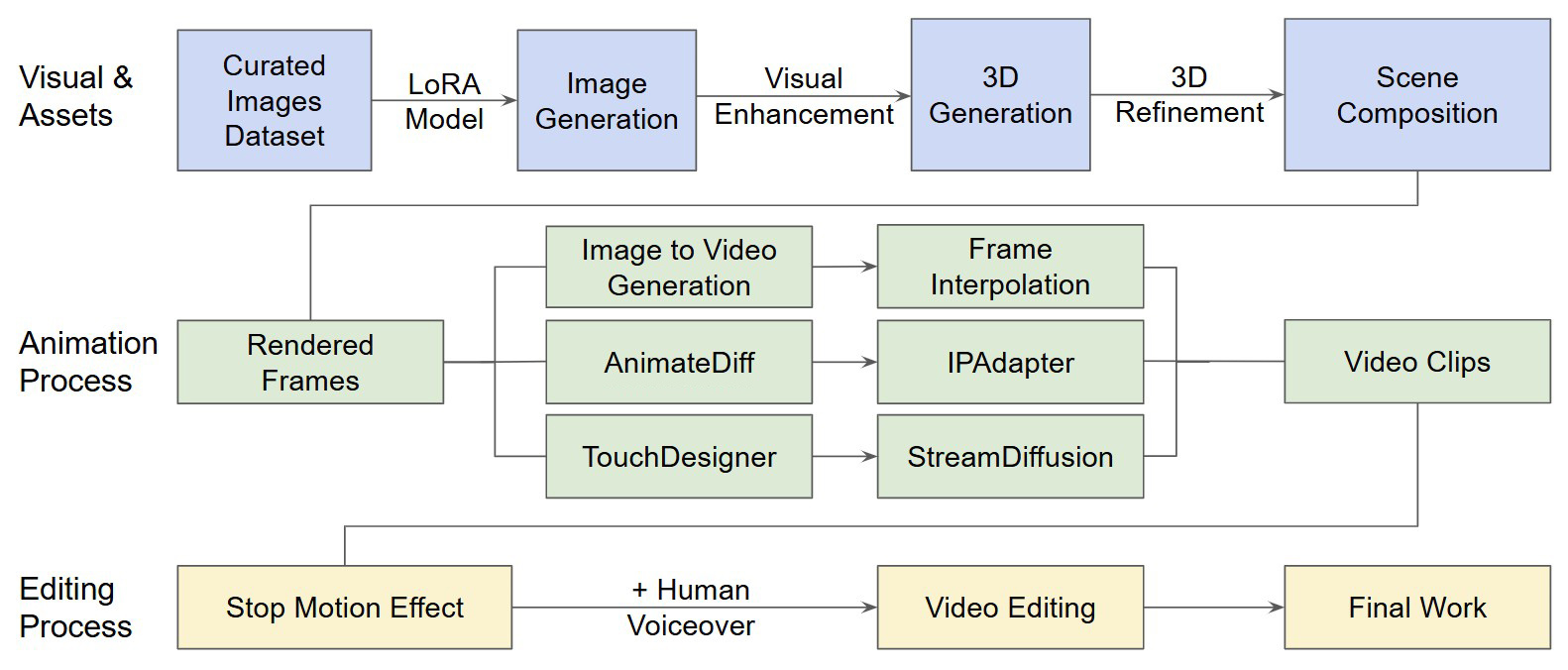}
    \caption{\textit{Overthinking}\cite{Overthinking} workflow\cite{hu2021lora}\cite{guo2023animatediff}\cite{liu2023ipadapter}\cite{touchdesigner}\cite{kodaira2023streamdiffusion}}
    \label{fig:Overthinking workflow.jpg}
\end{figure}

\subsubsection*{Shot Composition and Camera Control}

% While generative models can produce impressive imagery, they often lack  continuity and control over shot composition, limiting their use in extended narratives. To mitigate these constraints, filmmakers adopt creative techniques such as frame interpolation, start-end frame continuity, and hybrid 3D pipelines.
Generative models can produce striking visuals but often struggle with temporal continuity and compositional control, hindering their utility for extended narratives. Filmmakers mitigate these limitations by employing techniques such as frame interpolation, enforcing start-end frame consistency, and utilizing hybrid 3D pipelines.

%ControlNet guides Stable Diffusion effectively assists in shot composition\cite{zhang2023adding}, and many AI tools already integrate camera control functionalities for creators to better control image and video generation\cite{klingai,liu2024sora}. However, the duration of generated videos is typically under 20 seconds, making jump cuts necessary and posing a challenge to create a continuous viewing experience.

The 'start and end frame' approach, facilitates consistent camera movement and scene transitions, supporting content generation beyond the initial frame and enabling continuous 'one-take' effects. \textit{Invisible Women}\cite{Invisible_Women_2025}, for instance, leveraged this approach by stitching AI-generated segments into seamless one-take sequences, achieving a distinct visual and narrative style. Another strategy utilizes 3D technology for precise camera control. \textit{Dancestry}\cite{Dancestry2025} illustrates this, using AI primarily to generate detailed 3D assets, rigs, and facial expressions, which were then animated and rendered in Blender\cite{blender} to achieve fine-grained control over camera paths.
% To address these limitations, artists have developed creative strategies. The "start and end frame" technique excels in camera movement processing and scene transitions, allowing for the display of content beyond the initial frame and enabling "one-take" effects. For example, \textit{Invisible Women}\cite{Invisible_Women_2025} stitched AI-generated segments to produce seamless one-take sequences, creating a unique visual and narrative style. Also, utilizing 3D technology to control the camera is also an effective method. For instance, in \textit{Dancestry}\cite{Dancestry2025}, AI was employed to assist in creating detailed 3D assets, movement rigs, and facial expressions. Animation and rendering were performed in Blender\cite{blender}, allowing for precise control of camera movements.

\subsubsection*{Dataset Bias Challenges}

% Dataset limitations often introduce biases that pose challenges in visual generation. The character associated with many careers are typically male, thus the AI generated images are also typically male. However, in the film   \textit{Invisible Women}\cite{Invisible_Women_2025} , authors aimed to highlights issues of gender bias, occupational algorithm bias, and data disparities in AI-generated imagery processes. They  optimizing their prompts and parameters to ensure a more authentic portrayal of female characters and challenge entrenched stereotypes.
Biases embedded in training datasets pose significant challenges for equitable visual generation. A common manifestation is occupational gender bias, where models often default to generating male figures for professions predominantly represented by men in the data. Addressing this directly, the film \textit{Invisible Women}\cite{Invisible_Women_2025} sought to expose these occupational stereotypes within AI generation. The filmmakers meticulously refined prompts and model parameters, striving for more authentic representations of women and actively challenging conventional stereotypes.

% 	Eg. what sets of keywords do they use; how do they keep the style consistent. 《sacred dance》《cat can dance》
% 	Eg. how to enrich details in the view?
% Discussion of the role of artistic direction in shaping GenAI's visual output
% Eg. Do people do reframing? Color grading? What is left to do after AI generation?
% What else is important for the visual?
% Eg. 角度的丰富程度, 运镜的控制, 片段时长

\subsection{Case Study of New Artistic Expression Forms} 
% Rather than copying traditional filmmaking, many AI filmmakers embrace the limitations of generative models to cultivate new forms of visual language and narrative structure. These projects highlight how randomness, imperfection, and hybrid techniques can serve as expressive assets.
Beyond replicating traditional filmmaking techniques, some AI filmmakers turn perceived limitations of generative models—such as randomness and visual imperfections—into novel tools, reframing these characteristics as valuable resources for artistic expression.

%While many AI filmmakers strive for visual consistency, some creators embrace the inherent limitations of generative AI to forge new artistic expressions.

\subsubsection*{Embracing Randomness as a Creative Advantage}

\begin{figure*}[tp]
    \centering
     \includegraphics[width=0.7\linewidth]{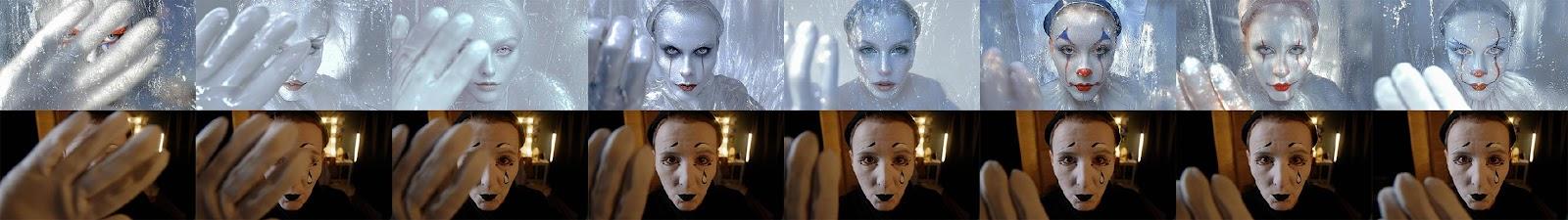}
    \caption{\textit{Clown}\cite{clown2025} frame by frame style transfer}
    \label{fig:clown.jpeg}
\end{figure*}

% The unpredictability of AI outputs can serve as a storytelling mechanism when aligned with thematic intent. \textit{CLOWN}\cite{clown2025} transformed Midjourney\cite{midjourney}'s random generation feature into a psychological storytelling tool. It employs a frame-by-frame stylization technique inspired by stop-motion animation. Each frame was individually processed through AI, maintaining visual continuity through a consistent art style while allowing subtle variations that mirror the protagonist's fragmented identity. This approach turned AI's inconsistency into a profound artistic statement about a clown gradually losing her sense of self (Fig \ref{fig:clown.jpeg}).
The unpredictability of AI outputs can become a storytelling device when aligned with thematic intent. \textit{CLOWN}\cite{clown2025} repurposed Midjourney\cite{midjourney}’s random generation feature as a psychological narrative tool, using a frame-by-frame stylization technique inspired by stop-motion animation. Each frame was individually processed through AI, maintaining visual coherence through a consistent art style while introducing subtle variations that reflect the protagonist’s fragmented identity. In doing so, the film transforms AI’s inconsistency into a poignant expression of a clown’s gradual loss of self (Fig.\ref{fig:clown.jpeg}).

\subsubsection*{Finding Poetry in the Imperfection of Image Generation}
% Visual imperfections in AI outputs — blurred edges, odd proportions, inconsistent lighting — can be repurposed as aesthetic features. \textit{Qatsi}\cite{qatsi} leverages abstraction to channel emotional and philosophical themes. The team adopted a monochromatic 4: 3 aspect ratio aesthetic and abstract narrative following Soviet montage theory\cite{eisenstein1923montage}. Inspired by the philosophy of the filmmaker David Lynch\cite{lynch2006catching}, the work embraces imperfections, irregularities, and errors, often seen as shortcomings, which became tools for artistic expression rather than obstacles. By allowing technology to guide them toward poetic expression, they discovered unexpected beauty emerging from imperfection.

Visual imperfections in AI outputs—blurred edges, odd proportions, inconsistent lighting—can be repurposed as aesthetic features. \textit{Qatsi}\cite{qatsi} employs abstraction to convey emotional and philosophical themes, adopting a monochromatic 4:3 aspect ratio and an abstract narrative structure rooted in Soviet montage theory\cite{eisenstein1923montage}. Inspired by the filmmaking philosophy of David Lynch\cite{lynch2006catching}, the project embraces the imperfections, irregularities, and errors in generated visuals—elements often seen as flaws—and reframes them as tools for artistic expression.
% By allowing technology to guide them toward poetic expression, they discovered unexpected beauty emerging from imperfection.

%\textit{Qatsi}\cite{qatsi} represents another approach. Rather than struggling against AI's limitations in photorealism, the creators embraced abstraction, the film embraced AI’s abstract, expressionistic potential to evoke emotion and introspection. The team adopted a monochromatic 4:3 aspect ratio aesthetic and abstract narrative following Soviet montage theory\cite{eisenstein1923montage}. Inspired by the philosophy of filmmaker David Lynch\cite{lynch2006catching}, the work embraces imperfections, irregularities, and errors—often seen as shortcomings—became tools for artistic expression rather than obstacles. By allowing technology to guide them toward poetic expression, they discovered unexpected beauty emerging from imperfection.

\subsubsection*{Reimagining Traditional Animation Techniques with AI}

% AI can replicate and augment traditional animation aesthetics by intentionally manipulating temporal resolution and image quality.  \textit{Overthinking}\cite{Overthinking} deliberately calibrated frame rates in After Effects to between 12-15 frames per second, both mitigating viewer discomfort from AI-generated motion artifacts and intentionally emulating stop-motion animation's distinctive aesthetic. Similarly, \textit{Round Table}\cite{Round_Table} merged AI-generated assets with traditional animation techniques, creating a unique workflow where AI-produced visuals were assembled in a stop-motion style, producing a handcrafted, tactile feel that countered the typically sleek appearance of AI imagery.

AI can replicate and enhance traditional animation aesthetics by deliberately manipulating temporal resolution and image quality. \textit{Overthinking}\cite{Overthinking} adjusted frame rates in After Effects to 12–15 frames per second, both reducing viewer discomfort from AI-generated motion artifacts and mimicking the distinctive look of stop-motion animation. Similarly, \textit{Round Table}\cite{Round_Table} combined AI-generated assets with traditional animation techniques, assembling the visuals in a stop-motion style to evoke a handcrafted, tactile quality that contrasts with the typically sleek appearance of AI imagery.

\subsubsection*{Experimental 3D Aesthetics}

% AI-generated 3D assets can have various forms of represtation , not only polygonal meshes but also volumetric representations like spatial splats representationf in   Gaussian splatting,paving ways for novel form of AI-based spatial storytelling..Films such as \textit{Metanoia}\cite{metanoia_mit_ai_film_2025} and \textit{Dressage Marching Through Memories}\cite{dressage_marching_memories_mit_ai_film_2025} turned the gaussian splats into  surreal and fluid visual spaces that defy conventional spatial logic\cite{kerbl3Dgaussians}, creating a mysterious and nostalgic feeling as this follow effect reminds people of unclear memory inside their hands.

AI-generated 3D assets can take diverse forms, extending beyond polygonal meshes to include volumetric representations such as spatial splats in Gaussian splatting—opening new possibilities for AI-driven spatial storytelling. Films like \textit{Metanoia}\cite{metanoia_mit_ai_film_2025} and \textit{Dressage Marching Through Memories}\cite{dressage_marching_memories_mit_ai_film_2025} leverage Gaussian splats\cite{kerbl3Dgaussians} to create surreal, fluid visual environments that evoke a nostalgic atmosphere, reminiscent of fragmented and fading memories filled with gaps and distortions.

 % that defy conventional spatial logic, evoking a mysterious, nostalgic atmosphere reminiscent of distant, half-remembered memories

%Beyond these approaches, films like \textit{Metanoia}\cite{metanoia_mit_ai_film_2025} and \textit{Dressage Marching Through Memories}\cite{dressage_marching_memories_mit_ai_film_2025} are beginning to explore advanced 3D technical innovations like Gaussian splatting—a technique that represents 3D scenes as a cloud of particles rather than polygonal meshes—opening new possibilities for visual representation through AI\cite{kerbl3Dgaussians}. These experimental approaches suggest that as generative AI tools continue to evolve, the most compelling artistic expressions may come from creators who embrace AI's unique characteristics to develop entirely new visual languages.

%  Identification of emerging art forms and genres that are uniquely enabled by GenAI.
% 	Eg. diffusion creates the illusion of morphing and dreaming. 《A dream about to awaken》
% 	Eg. nerf reconstruction. 《颅内花园》
% Eg. create unusual characters 《离谱村》
% Eg. Create abstract unworldly shape. 《Shaun Dougherty's 'Dance In Life' a AI DreamCore Musical in 4K》
% Eg. 生成难以拍摄的素材 《Qatsi》

\begin{figure}[tp]
    \centering
     \includegraphics[width=\linewidth]{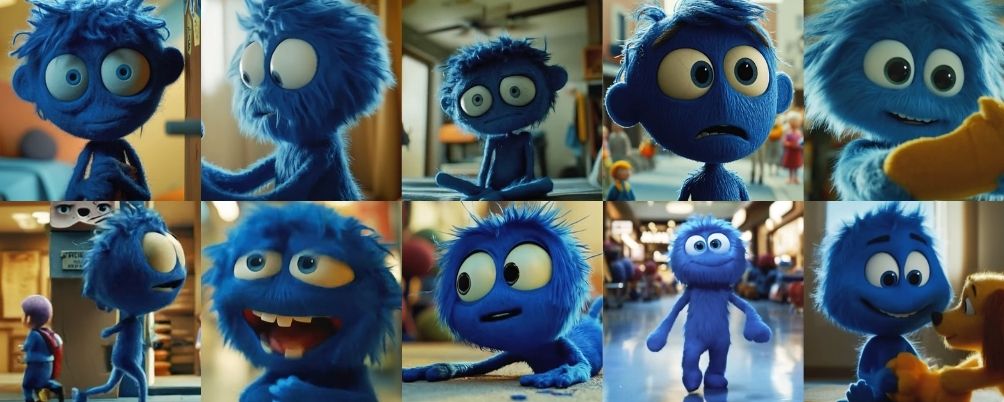}
    \caption{\textit{For Pixi}\cite{forpixi2025} character design and consistency control}
    \label{Pixi Character.jpg}
\end{figure}

\subsection{Case Study of Character Creation with GenAI}

Character design in AI filmmaking requires a balance of consistency, emotional depth, and technical flexibility. Although generative tools allow for rapid prototyping, artists must maintain continuity and expressiveness through iterative refinement and careful human intervention.
%Character creation is central to AI-assisted filmmaking, defining both the narrative and visual identities of AI-generated works. Among 118 film submissions in 2025, 109 featured characters, with 77 relying solely on digital characters, 16 blending real and digital elements, and 16 including only real actors.

\subsubsection*{Character Design and Visual Consistency}
% AI enables imaginative character design through the fusion of disparate concepts, but maintaining consistency across scenes requires intentional prompting and manual editing. In \textit{Tale of Lipu Village}\cite{tale_of_lipu_village}, characters such as broccoli-growing sheep and fried-egg flowers reflect AI's capacity to blend unrelated forms, resulting in cohesive aesthetics otherwise difficult to achieve via traditional methods. \textit{Dance of E-Spark}\cite{dance_of_e_spark_mit_ai_film_2025} used style-specific prompts and Photoshop\cite{adobe_photoshop} adjustments to ensure consistency in robotic character design.
AI enables imaginative character design by blending disparate concepts, but maintaining consistency across scenes demands deliberate prompting and manual refinement. In \textit{Tale of Lipu Village}\cite{tale_of_lipu_village}, characters like broccoli-growing sheep and fried-egg flowers showcase AI's ability to merge unrelated elements into a cohesive aesthetic that would be challenging to achieve through traditional methods. \textit{Dance of E-Spark}\cite{dance_of_e_spark_mit_ai_film_2025} achieved visual consistency in its robotic characters through style-specific prompts and post-processing using Photoshop\cite{adobe_photoshop}.

%Films like \textit{Tale of Lipu Village}\cite{tale_of_lipu_village} demonstrate how generative AI can merge disparate concepts—e.g., flowers shaped like fried eggs, sheep that grow broccoli—to yield imaginative character designs. Such unlikely combinations exploit AI’s tendency to blend unrelated forms, resulting in cohesive aesthetics otherwise difficult to achieve via traditional methods. In \textit{Dance of E-Spark}\cite{dance_of_e_spark_mit_ai_film_2025}, the production team used style-specific prompts to create robotic characters, refining their lighting and poses in Photoshop. These examples underscore GenAI’s capacity to produce distinctive visuals through informed prompting and supplemental manual edits.

% Maintaining consistent character features across multiple scenes is especially challenging in AI-generated films. \textit{For Pixi}\cite{forpixi2025} illustrated the importance of iterative refinement and prompt engineering by using the same prompt—“A claymation, puppet-style 3D animation world”—in every Midjourney\cite{midjourney} generation. Although subtle inconsistencies such as differing eye shapes persisted, tools like Midjourney’s region-specific editor and 'vary' function proved invaluable. By generating 30–40 variations per prompt iteration and applying negative prompts to exclude unwanted elements, creators minimized undesirable discrepancies and sustained character uniformity throughout the film (Fig \ref{Pixi Character.jpg}).

Maintaining consistent character features across multiple scenes is especially challenging in AI-generated films. \textit{For Pixi}\cite{forpixi2025} illustrated the importance of iterative refinement and prompt engineering by using the same prompt—“A claymation, puppet-style 3D animation world”—in every Midjourney\cite{midjourney} generation. Although subtle inconsistencies such as differing eye shapes persisted, tools like Midjourney’s region-specific editing tools proved invaluable. By generating 30–40 variations per prompt iteration and applying negative prompts to exclude unwanted elements, creators minimized undesirable discrepancies and sustained character uniformity throughout the film (Fig \ref{Pixi Character.jpg}).

\subsubsection*{Character Motion and Emotion}

Convincing motion and emotional expression remain among the most technically demanding aspects of AI character creation. Approaches range from text-based prompts that describe movement to live-action motion capture. In \textit{O.R.V. 8 Oscillating Rhythmic Vinyl}\cite{orv8_mit_ai_film_2025}, a simple green screen setup allowed the performance of an actor captured on camera to be transformed into a fully animated robot using Wonder Studio\cite{wonder_studio}. The resulting sequences were composited into AI-generated environments in Adobe After Effects\cite{adobe_after_effects} and Adobe Premiere Pro\cite{adobe_premiere_pro}, providing a seamless mix of human expressiveness and AI-driven aesthetics. 

% Believable emotional expression is vital for AI-created characters. \textit{The Last Dance}\cite{lastdance2025} employed emotion-guided animation, where the film prompts specified how characters should feel and move in each scene.
Touching emotional expression is essential for AI-generated characters. \textit{The Last Dance}\cite{lastdance2025} employed emotion-guided animation, using prompts to articulate how those emotions should be physically expressed through movement in each scene.

% \subsubsection*{Character Voiceover}

% While AI voice synthesis tools offer customization, they often lack emotional nuance. Some filmmakers supplemented AI dubbing with human voiceover. Others, like \textit{Synthetic Rhythm}\cite{SYNTHETIC_RHYTHM_2025}, ventured deeper into customizable voice-synthesis models. By carefully tuning pitch, tone, and pacing, and weaving these enhanced voices into the film’s sound track, it created a captivating and authentic character voice.

%In summary, AI-assisted character creation benefits from an array of strategies—prompt-driven design, meticulous consistency checks, hybrid motion capture, emotion-driven animation, and finely tuned voice models. By balancing AI’s capacity for rapid, inventive output with manual refinement and human-led workflows, filmmakers can produce characters that are visually distinct, coherent across scenes, and emotionally compelling on screen.

% A brief mention of what percentage of the film contains character. 
% Eg. xxx films contain a consistent character. 
% Strategies and techniques for maintaining character consistency across various GenAI outputs
% Eg. Conditioned on image of characters, have detailed text descriptions (examples).
% Exploration of GenAI's potential in character movement and facial expression.
% Eg. Interview the tricks from 
% 《Shaun Dougherty's 'Dance In Life' a AI DreamCore Musical in 4K》
% 《Mr Fox part 2》
% Tricks for creating characters in the genAI era.
% Eg. 特写腿，手，背影，某些feature保持一致就好了 雀斑 《Endless night》

\begin{figure*}[tp]
    \centering
     \includegraphics[width=0.6\linewidth]{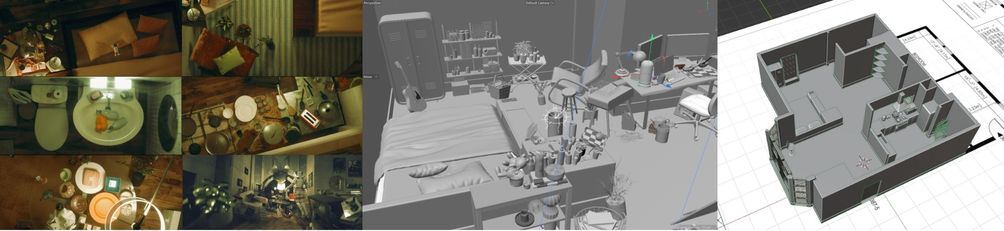}
    \caption{\textit{Fish Tank}\cite{The_Fish_Tank} final scenes, 3D assets and models}
    \label{Fish Tank.jpg}
\end{figure*}

\subsection{Case Study of 3D Generation in Filmmaking}

% AI-based 3D generation streamlines asset creation but still requires human intervention for quality control, rigging, and stylistic coherence. The current genAI tools generates assets much faster than humansbut faces limitations in topology, animation readiness, and abstract design flexibility.
AI-based 3D generation is significantly faster than manual modeling but still requires human input for quality control, rigging, and stylistic coherence. Current tools struggle with topology, animation readiness, and abstract design flexibility.

%AI-assisted 3D modeling provides filmmakers with powerful tools for creating, integrating, and animating digital assets and scenes more efficiently than traditional 3D pipelines, as well as special effects.

\subsubsection*{Mixing Real Footage and 3D Content}

% Blending AI-generated 3D characters into live-action environments requires precise control of lighting, depth, and perspective. Films like \textit{CLOWN}\cite{clown2025} employed Wonder Studio\cite{wonder_studio} , whose AI pipeline automatically animates, lights, and composes CG characters into real-world scenes. By adjusting lighting and depth, the film\cite{clown2025} ensures that digital character align with live-action camera angles and environments, preserving the authenticity of the hybrid sequences.

Blending AI-generated 3D characters into live-action environments requires precise control of lighting, depth, and perspective. Films like \textit{CLOWN}\cite{clown2025} used Wonder Studio\cite{wonder_studio}, an AI pipeline that automates animation, lighting, and compositing of CG characters into real-world scenes. By fine-tuning lighting and depth, Wonder Studio ensures digital characters align seamlessly with live-action camera angles and settings, preserving the authenticity of hybrid sequences\cite{wonder_studio}.

\subsubsection*{3D Asset and Scene Generation}

% Tools such as Luma AI\cite{luma_ai}, Meshy\cite{meshy}, and Hyper3D\cite{hyper3d} significantly accelerate asset creation workflows. In \textit{Fish Tank}\cite{The_Fish_Tank}, filmmakers tested Photo-to-3D, Text-to-3D, and AI Image-to-3D pipelines, reducing asset modeling time by over 90\%, producing a model in less than 5 minutes rather than 1–2 hours of meticulous sculpting. Luma AI\cite{luma_ai} captures real-world environments as high-fidelity and 3D meshes for scene creation. And Meshy\cite{meshy} quickly transforms prompts or 2D images into workable 3D models.

Tools like Luma AI\cite{luma_ai}, Meshy\cite{meshy}, and Hyper3D\cite{hyper3d} drastically speed up asset creation. In \textit{Fish Tank}\cite{The_Fish_Tank}, filmmakers tested Photo-to-3D, Text-to-3D, and Image-to-3D pipelines, cutting modeling time by over 90\%—from 1–2 hours to under 5 minutes. Some tools capture real-world environments as high-fidelity 3D meshes, while other tools convert prompts or 2D images into usable 3D models within minutes.

%\textit{Fish Tank}\cite{The_Fish_Tank}, an experimental short, illustrates how AI drastically reduces the labor of 3D asset creation, but also reveals current limitations. Tools such as Luma AI\cite{luma_ai}, Meshy\cite{meshy} and Hyper3d\cite{hyper3d} cut asset generation time by more than 90\%, producing a model in less than 5 minutes rather than 1–2 hours of meticulous sculpting. Luma AI\cite{luma_ai} captures real-world environments as high-fidelity and 3D meshes for scene creation. And Meshy\cite{meshy} quickly transforms prompts or 2D images into workable 3D models. In \textit{Fish Tank}\cite{The_Fish_Tank}, the team used three distinct workflows: Photo-to-3D, AI-generated Image-to-3D, and Text-to-3D, to expedite production and concentrate on storytelling.

However, certain AI limitations persisted in the production of \textit{Fish Tank}\cite{The_Fish_Tank}. First, UV Mapping and Material Application: Meshy\cite{meshy}’s initial UV mapping capabilities were weak, requiring additional manual adjustments for accurate texturing. Later improvements—particularly the introduction of quad-based topology—enhanced asset usability but still required refinement. Second, AI-generated mesh structures often lacked animation-ready topology. The rigid, mechanical nature of the generated models made them unsuitable for skeletal rigging and deformation without manual retopology. Third, AI struggled to generate complex, abstract objects with high creative flexibility, often defaulting to standardized geometric forms. This constrained its application to highly conceptual or surreal scenes (Fig. \ref{Fish Tank.jpg}).

% AI-based 3D generation continues to evolve, promising enhanced UV mapping, automated topology, and improved stylistic consistency. As these capabilities advance, AI will play an increasingly central role in lowering production costs, accelerating workflows, and empowering a broader range of creators to produce high-caliber 3D content.

% 《Dancestry》
% Overview of the 3D generation pipeline in the context of GenAI
% Eg. generate the mesh, and do coloring later or object directly?
% Eg. generate the objects or characters?
% What is needed after the 3D generation?
% Eg. how to animate the objects? 《Dancestry》
% Eg. how to keep the generated objects consistent with the environment?
% Mixing Real Shot and Generated Content
% Techniques for seamlessly blending real footage with GenAI-generated elements.
% Eg. Wonder studio 《CLOWN》《Through The Infinity》
% How does AI work together with various kinds of shots?
% Eg. framing, angle.

\begin{figure*}[tp]
    \centering
     \includegraphics[width=0.65\linewidth]{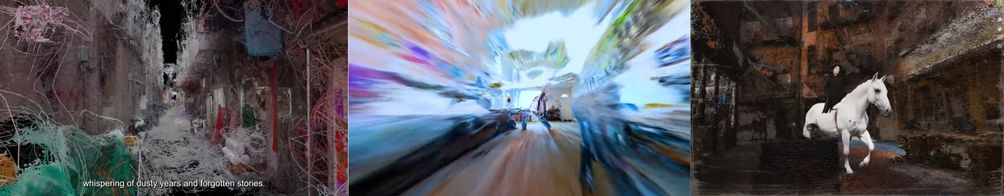}
    \caption{Volumography scenes in \textit{Former Garden}\cite{FORMER_GARDEN_2024}, \textit{Metanoia}\cite{metanoia_mit_ai_film_2025} and \textit{Dressage Marching Through Memories}\cite{dressage_marching_memories_mit_ai_film_2025}}
    \label{AI Volumography.jpg}
\end{figure*}

\subsection{Case Study of AI in XR Filmmaking}

AI is reshaping XR filmmaking through real-time compositing, immersive camera control, and volumetric storytelling. These advances expand the director’s creative toolkit and blur the line between cinematic and interactive experiences.
%AI is rapidly reshaping Extended Reality (XR) filmmaking through two key advancements: AI-driven immersive videos and AI volumography.

\subsubsection*{AI-Assisted Immersive Videos}

AI enhances immersive filmmaking through motion capture automation, scene generation, and real-time compositing. In \textit{Machine Learning}\cite{machine_learning_mit_ai_film_2025}, motion-captured animations were composited into 180° stereoscopic videos using AI tools such as Wonder Studio\cite{wonder_studio}, which captured intricate body and hand movements for seamless integration. Additionally, Project Reframe\cite{autodesk2024reframe} enabled the use of headset cameras to track hand gestures, expanding interaction fidelity within VR environments.

%Immersive videos gain depth and interactivity from automated scene generation, motion capture, and real-time compositing. Tools like Wonder Studio\cite{wonder_studio} facilitate AI-driven motion capture by allowing creators to accurately capture intricate hand and body movements, which can then be blended into 360° footage for a more adaptive, responsive user experience. For instance, the \textit{Machine Learning}\cite{machine_learning_mit_ai_film_2025} project uses AI-assisted compositing to integrate motion-captured animations directly onto 180° 3D video, creating a seamless XR environment. Project Reframe\cite{autodesk2024reframe} was also used to capture hand movements for animation with headset cameras. The final film could be viewed in VR Headset.

\subsubsection*{AI Volumography}

Volumetric filmmaking—powered by techniques such as NeRF\cite{mildenhall2021nerf}, Gaussian splatting\cite{kerbl3Dgaussians}, and point cloud rendering—redefines cinematic authorship by decoupling scene capture from camera control. Unlike traditional filmmaking, which depends on fixed, discrete frames, volumography records entire scenes as dynamic, navigable 3D or 4D datasets.

% Volumetric filmmaking—enabled by NeRF\cite{mildenhall2021nerf}, Gaussian splatting\cite{kerbl3Dgaussians}, and point cloud rendering—fundamentally redefines cinematic authorship by separating capture from camera control. Unlike traditional filmmaking’s reliance on discrete frames, volumography records entire scenes as dynamic, navigable 3D / 4D datasets.

%Volumography—powered by neural radiance field (NeRF) rendering, Gaussian splatting, photogrammetry, and volumetric video—goes beyond typical 2D or 3D video captures\cite{kerbl3Dgaussians,martin2021nerf,mildenhall2021nerf}. Unlike traditional filmmaking’s reliance on discrete frames, volumography records entire scenes as dynamic, navigable 3D / 4D datasets.

\textit{Metanoia}\cite{metanoia_mit_ai_film_2025}, for example, captured a dancer using a NeRF-based pipeline and later explored a wide range of camera movements in post-production, reversing the usual order of shot planning. This non-linear process grants directors the freedom to experiment with pacing, composition, and viewpoint well after principal capture. Similarly, \textit{Former Garden}\cite{FORMER_GARDEN_2024} demonstrates volumography’s capacity for introspective storytelling, using a point cloud representation of fragmented memories to depict the protagonist’s subconscious (Fig. \ref{AI Volumography.jpg}).

Volumography integrates spatial computing with cinematic storytelling, offering filmmakers unprecedented flexibility through unlimited reshoots, unrestricted camera movements, and seamless real-CG integration, thereby revolutionizing immersive and interactive narrative experiences in XR environments.

%By merging spatial computing with cinematic storytelling, volumography enables unlimited reshoots, unrestricted camera paths, and seamless real-CG integration in XR environments. Whether it’s for immersive dance performances, urban documentation, or deeply personal narratives, these AI-driven approaches grant filmmakers new levels of flexibility and creative control. As a result, volumography is poised to become a cornerstone of future visual media, redefining how audiences engage with narrative spaces and making complex, interactive storytelling more accessible than ever.

\begin{table*}[htbp]
\centering
\begin{tabular}{|p{2.2cm}|p{3.3cm}|p{2.8cm}|p{3.7cm}|p{3cm}|}
\hline
\textbf{Pipeline Type} & \textbf{Typical Films} & \textbf{Typical Tools} & \textbf{Advantages} & \textbf{Challenges} \\
\hline
2D AI Pipeline (Text-Image-Video) & \textit{For Pixi}\cite{forpixi2025} \newline \textit{Qatsi}\cite{qatsi} \newline \textit{Sacred Dance}\cite{sacreddance2025} & OpenArt\cite{openai2024} \newline Midjourney\cite{midjourney} \newline Pixverse\cite{pixverse} & Quick iteration; strong visual style control; accessible tools & limited motion control; short video duration \\
\hline
3D Generation Pipeline & \textit{Dancestry}\cite{Dancestry2025} \newline \textit{Overthinking}\cite{Overthinking} & Meshy\cite{meshy} \newline Blender\cite{blender} \newline Hyper3D\cite{hyper3d} \newline LumaAI\cite{luma_ai} & Accurate modelling and camera control & Complex workflow; time-intensive \\
\hline
Hybrid Live Action + AI & \textit{Metanoia}\cite{metanoia_mit_ai_film_2025} \newline \textit{Clown}\cite{clown2025} & Wonder Studio\cite{wonder_studio} \newline Touchdesigner\cite{touchdesigner} & Emotive performance capture; style remix; strong narrative flexibility & Integration complexity; lighting and depth mismatch \\
\hline
XR / Volumetric Pipeline & \textit{Former Garden}\cite{FORMER_GARDEN_2024} \newline \textit{Metanoia}\cite{metanoia_mit_ai_film_2025} \newline \textit{Machine Learning}\cite{machine_learning_mit_ai_film_2025} & NeRF\cite{martin2021nerf,mildenhall2021nerf} \newline Unity3D\cite{unity} \newline Unreal Engine\cite{unreal_engine}  & immersive and interactive potential, post-capture camera control & High computational cost; dataset limitations \\
\hline
\end{tabular}
\caption{Overview of AI filmmaking pipelines, representative works, typical tools, benefits, and associated challenges.}
\label{tab:ai_pipelines}
\end{table*}

\subsection{Case Study of Creative Tech in Filmmaking} 

AI also enables new creative workflows in filmmaking by merging human vision with generative AI capabilities.

% The most compelling AI-driven films emerge from hybrid workflows that combine human vision with GenAI Tools. These projects show how AI not only accelerates production but also inspires new narrative structures, aesthetic expressions, and collaborative modalities.

%AI-driven filmmaking is not only about accelerating production pipelines but also about enabling novel, unconventional creative workflows that redefine how films are conceived and crafted. Filmmakers are pushing the boundaries of AI technologies to achieve unique artistic expressions.

\begin{figure*}[!htbp]
    \centering
    \includegraphics[width=0.65\linewidth]{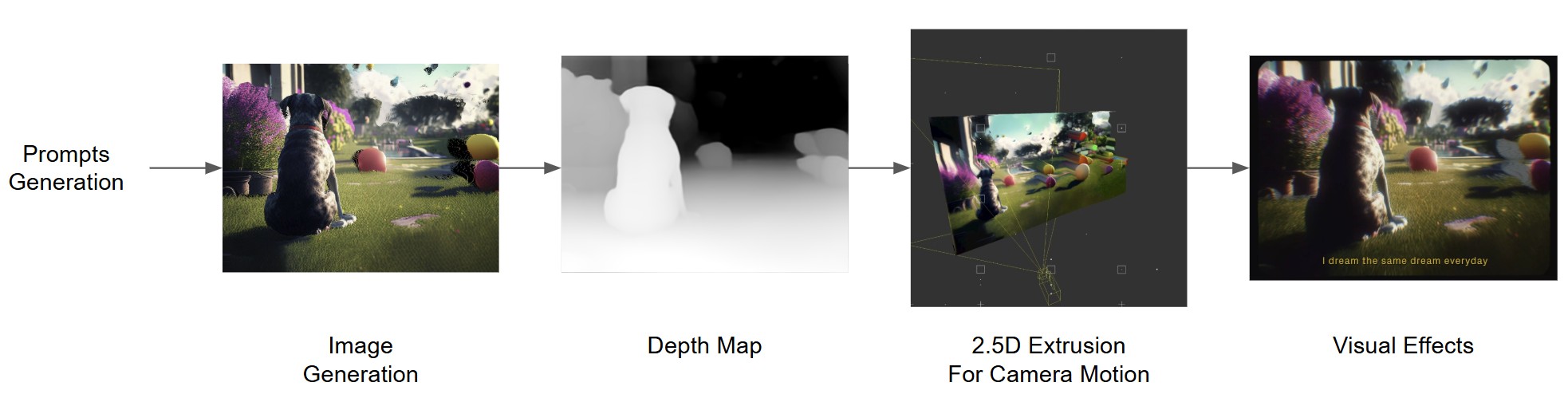}
    \caption{\textit{DOG: Dream of Galaxy}\cite{DOG_2023} workflow}
    \label{fig:DOG_Workflow}
\end{figure*}

\subsubsection*{Human-Machine Collaborative Approach}

AI becomes most powerful when used as a collaborator rather than an autonomous system. 2023 Best Film Winner \textit{DOG: Dream of Galaxy}\cite{DOG_2023} was created long prior to the emergence of advanced AI video generation tools. The production began with script-driven prompt engineering to ensure narrative coherence. The Midjourney\cite{midjourney}-generated images were processed through Stable Diffusion\cite{thygate_depthmap_script} to create depth maps. These depth-enhanced images were imported into Cinema 4D\cite{cinema_4d} and extruded into 2.5D models, allowing precise control over camera movement, focal depth, and composition. AI-generated voices were enhanced with manual reverb effects, synchronized with electronic sound cues mimicking mechanical operations. This innovative workflow bridges AI-generated content with traditional filmmaking techniques, showcasing the potential of human-machine collaboration in cinematic expression.

% The film adopted a 4:3 aspect ratio, reminiscent of 1980s sci-fi films. Analog-inspired imperfections, such as film grain and scratches, were added in After Effects\cite{adobe_after_effects} to reinforce the nostalgic aesthetic. 

\subsubsection*{Hybrid Live Action, 2D and 3D Workflow}
% Integrating generative tools with live-action footage allows filmmakers to juxtapose the organic and synthetic within a unified visual logic. Set in a dystopian world where humanity is enslaved by digital masks, \textit{Metanoia}\cite{metanoia_mit_ai_film_2025}  juxtaposes cold, algorithmic precision with the raw emotion of dance. The protagonist’s struggle to reclaim her humanity unfolds through a hybrid workflow:
% Real-Time AI Integration: TouchDesigner\cite{touchdesigner} and Stable Diffusion\cite{Stable_Diffusion_2022_CVPR} dynamically generate abstract visual textures, blending them with live-action footage to symbolize the encroachment of AI into her sanctuary. Luma AI\cite{luma_ai} captured dancers’ movements in 3D, which were later distorted using AI effects to create fluid, surreal transformations. This fusion of physical performance and digital abstraction elevates dance as a metaphor for rebellion.
Integrating generative tools with live-action footage allows filmmakers to juxtapose the organic and synthetic within a unified visual logic. Set in a dystopian world where humanity is enslaved by digital masks, \textit{Metanoia}\cite{metanoia_mit_ai_film_2025}  juxtaposes cold, algorithmic precision with the raw emotion of dance. The protagonist’s struggle to reclaim her humanity unfolds through real-time AI integration: TouchDesigner\cite{touchdesigner} and Stable Diffusion\cite{Stable_Diffusion_2022_CVPR} dynamically generate abstract visual textures, blending them with live-action footage to symbolize the encroachment of AI into her sanctuary. Luma AI\cite{luma_ai} captured dancers’ movements in 3D, which were later distorted using AI effects to create fluid, surreal transformations. This fusion of physical performance and digital abstraction elevates dance as a metaphor for rebellion.

\subsubsection*{Nonlinear Editing of AI-Generated Visuals and Music}
% AI also enables new temporal strategies in post-production, particularly when sound and visuals are developed in parallel. In \textit{For Pixi}\cite{forpixi2025}, nonlinear editing techniques were employed to synchronize AI-generated music and soundscapes with the evolving visuals. Rather than scoring music to a locked edit, the AI-generated music was produced iteratively alongside visual development, going back-and-forth between Premiere Pro\cite{adobe_premiere_pro} and Ableton Live\cite{ableton_live}. This interplay between   sound and image genaration led to a more cohesive, emotionally resonant audiovisual experience. The flexibility of AI-generated music allowed for continuous adjustments, enabling the filmmakers to refine the narrative rhythm dynamically during post-production.
AI also enables new temporal strategies in post-production, particularly when sound and visuals are developed in parallel. In \textit{For Pixi}\cite{forpixi2025}, nonlinear editing techniques were used to synchronize AI-generated music and soundscapes with the evolving visuals. Rather than scoring music to a locked edit, the soundtrack was developed iteratively alongside visual development, with frequent back-and-forth between Premiere Pro\cite{adobe_premiere_pro} and Ableton Live\cite{ableton_live}. This dynamic interplay between sound and image generation resulted in a more cohesive and emotionally resonant audiovisual experience. The flexibility of AI-generated music allowed for ongoing adjustments, enabling filmmakers to refine the narrative rhythm throughout post-production.
% These examples underscore how creative uses of AI technology extend beyond automation, opening new aesthetic possibilities and redefining artistic workflows in contemporary filmmaking.

% !TEX root = ../main.tex
%\section{Discussion}

%The case studies highlight both the vast potential and current limitations of AI-assisted filmmaking. While AI technologies have opened up unprecedented creative opportunities, ranging from character design to immersive volumetric storytelling, they also introduce challenges that require thoughtful human intervention. Based on the analysis across various domains including visual styling, character creation, 3D asset generation, motion, and sound, several key insights and recommended pipelines emerge for filmmakers seeking to adopt AI in their creative process (Table~\ref{tab:ai_pipelines}).

% !TEX root = ../main.tex
\section{Conclusion}
% \label{sec:conclusion} @ruihan

% Generative AI is reshaping filmmaking, significantly streamlining traditional workflows while simultaneously expanding the scope of creative storytelling. This survey reviewed recent advances and challenges in AI film creation, highlighting growing adoption in character animation, stylistic coherence, and immersive storytelling. While artists increasingly rely on generative tools for rapid asset generation and innovative visual styles, key challenges remain—particularly regarding consistency in character portrayal, nuanced motion control, and effective integration with real-world footage.

In summary, this paper surveys key developments in using generative AI for film creation, examining its impact on tasks like character animation, aesthetic styling, and 3D asset generation. (Table~\ref{tab:ai_pipelines}). While AI workflows reduce production costs and expand creative experimentation, they also pose challenges—particularly around character consistency, nuanced motion control, and blending AI outputs with real footage. Looking ahead, as generative AI tools advance and creative communities adapt, artists and AI will converge on integrated workflows that yield increasingly boundary-pushing cinematic experiences.

\clearpage
{
    \small
    \bibliographystyle{ieeenat_fullname}
    \bibliography{main}
}

\newpage

\section*{Appendix}

\section*{Acknowledgements}

We thank the organizers of the MIT AI Filmmaking Hackathons for providing a platform that fosters interdisciplinary exploration at the intersection of generative AI and cinematic storytelling. We are especially grateful to the participating filmmakers for generously sharing their creative processes, tools, and insights, which made these case studies possible, especially the winning-film creators, including Yidi Zhou, Prisha Jain, Olivia Lee from \textit{Qatsi} \cite{qatsi}; Leah Jiaxin Yu, Nate Zucker, Carolina Herrera from \textit{For Pixi} \cite{forpixi2025}; Yifei Li, Qianhui Sun, Zimeng Luo, Yongqi Liang from \textit{Dance of E-Spark}~\cite{dance_of_e_spark_mit_ai_film_2025};  
Yunlei Liu, Yujia Huang, Yiqi Li, Xiaoxu Han, Haokun Feng from \textit{Metanoia}~\cite{metanoia_mit_ai_film_2025}; Anton, Alexia, Caroline from \textit{Machine Learning}~\cite{machine_learning_mit_ai_film_2025}; She Yufeng, Yang Shuqi, Penelope, Qingcheng, Liguo, Damon from \textit{A Dream About to Awaken} \cite{dream2025}; Federico Agudelo from \textit{O.R.V. 8}~\cite{orv8_mit_ai_film_2025}; Qihan Jiang, Haoren Zhong, Beatrice Mai, Zongshuai Zhang from \textit{Round Table} \cite{Round_Table} ;
Song Lu, Fan Yu from \textit{Invisible Women} \cite{Invisible_Women_2025} ; Ellen Pan, Chris McLaughlin, Trinity Dysis, Justin Donovan, Josh Usita from \textit{Dancestry} \cite{Dancestry2025}; Haixin Yin, Zihao Zhang, Jianuo Xuan, Shengtao Shen, Fei Deng from \textit{Dressage Marching Through Memories}~\cite{dressage_marching_memories_mit_ai_film_2025}; Yiwei Xie, Minyu Chen from \textit{The Last Dance} \cite{lastdance2025}; Roos van der Jagt from \textit{Synthetic Rhythm}~\cite{SYNTHETIC_RHYTHM_2025}; Muwen Li, Xuanxuan Liu from \textit{Fish Tank}~\cite{The_Fish_Tank}; AJ, Gerry Huang, Shuai Shao, AI Yiran, Elfe Xu from \textit{Tale of Lipu Village}~\cite{tale_of_lipu_village}, Xiangning Yan, Hao Yu, Zhiyuan Zhou, Haowen Huang, Runtian Yang from \textit{Former Garden}~\cite{FORMER_GARDEN_2024}, Liu Yang, Candice Wu from \textit{DOG: Dream of Galaxy} \cite{DOG_2023}.

\end{document}